# Constructing a Testbed for Psychometric Natural Language Processing

Ahmed Abbasi[1], David G. Dobolyi[1], Richard G. Netemeyer[2]

[1]University of Notre Dame, Notre Dame, Indiana, USA
[2]University of Virginia, Charlottesville, Virginia, USA
E-mail: aabbasi@nd.edu, ddobolyi@nd.edu, rgn3p@virginia.edu

**Abstract**

Psychometric measures of ability, attitudes, perceptions, and beliefs are crucial for understanding user behaviors in various contexts including health, security, e-commerce, and finance. Traditionally, psychometric dimensions have been measured and collected using survey-based methods. Inferring such constructs from user-generated text could afford opportunities for timely, unobtrusive, collection and analysis. In this paper we describe our efforts to construct a corpus for psychometric natural language processing (NLP). We discuss our multi-step process to align user text with their survey-based response items and provide an overview of the resulting test bed which encompasses survey-based psychometric measures and accompanying user-generated text from over 8,500 respondents. We report preliminary results on use of the text to categorize/predict users' survey response labels. We also discuss the important implications of our work and resulting testbed for future psychometric NLP research.

## 1. Introduction

Psychometrics is the field of study concerned with the measurement of individuals' knowledge, abilities, attitudes, personality traits, and perceptions (Rust and Golomobok 2014). In social science research, psychometric dimensions are latent constructs that are known to be important antecedents, moderators, mediators, and consequents for important humanistic behaviors and outcomes (Li et al. 2016; 2017; 2020). For example, in the security behavior literature, constructs such as threat severity and response efficacy of protective mechanisms are critical psychometric measures of one's likelihood to avoid security threats (Dobolyi et al. 2017; Zahedi et al. 2015). In behavioral health, psychometric dimensions such as health numeracy, subjective health literacy, trust in physicians, and anxiety visiting the doctor's office are known to effect various health and wellness outcomes such as future physician visits and all-around well-being (Netemeyer et al. 2019). In electronic commerce, satisfaction with a website's functional, information, and visual design are correlated with purchase propensity and customer loyalty (Cyr 2009). Similarly, many individualized financial behaviors can be partially explained by financial literacy and psychological traits (Fernandes et al. 2014).

Given the importance of psychometric dimensions for understanding behaviors and outcomes in various domains, rigorous data collection protocols and best practices have been developed over the years (Netemeyer et al 2003). The primary modes of collection involve surveys and interviews (Li et al. 2020). While these techniques afford many benefits such as measurement control and robustness checks, they are not without their limitations. First, primary data collection facilitated through an administered survey can be time-consuming and invasive (often requiring 20-30 minutes of the respondents' time and attention). Second, such primary data collection cannot occur in real-time. Most surveys in field studies are conducted periodically at monthly or quarterly intervals. Third, while surveys are a rigorous form of data collection, they are limited in their ability to account for data/observations outside the pre-defined measurement framework. Consequently, effectively collecting and measuring relevant psychometric dimensions in a timely, unobtrusive, and more open-ended manner could be invaluable in many real-world settings (Gefen and Larsen 2017). Accurately and efficiently measuring psychometrics inherent in secondary data such as user-generated context may constitute an important complementary information access refinement to periodic survey-based data collection, with positive implications for information retrieval, mobile text analytics, and behavior modeling (Abbasi et al. 2015; Brown et al. 2015; Dong et al. 2019).

Given the abundance of user-generated text content, with unstructured data accounting for more than 80% of all data in various organizations (Kuechler 2007), measuring psychometric dimensions from text using natural language processing methods seems like one worthwhile avenue of inquiry (Li et al. 2018; 2020). In this paper we describe our efforts to construct a testbed for psychometric natural language processing (NLP). In the same vein as prior work on constructing language resources for sentiment, emotion, affect, and personality traits (Wiebe et al. 2005; Thelwall eta l. 2010; Luyckx and Daelemans 2008), we describe our approach and resulting test bed related to psychometric dimensions such as trust, anxiety, literacy, and numeracy in the health context. Figure 1 presents a motivating example describing the goal of our work. Given a well-established survey-based scale for "trust in visiting the physician's office," how can we arrive at a similar score for a user based on their generated text?

Any NLP-based approximation is likely to have measurement error due to the error of the text classifier trained to score the user text, as well as dissonance between a user's survey responses and text utterances. Nevertheless, the hope is that the ability to infer an imperfect yet reasonably accurate NLP-based measurement can still be advantageous as an alternative, complementary measure that can be derived unobtrusively in near real-time.

**Figure 1**: Illustration of survey and text data for a given psychometric dimension: "Trust in a Physician"

In this paper we describe the process taken to construct a psychometric NLP testbed. The testbed is comprised of user-generated text from over 8,000 individuals related to four key health-related psychometric dimensions of interest: trust in physicians, anxiety visiting the doctor's office, health numeracy, subjective health literacy. We believe our construction method and resulting test bed contribute to the language resource literature in the following ways:

- While psychometric dimensions such as sentiment, emotion, affect, and personality traits have garnered a fair amount of attention from the NLP community (Hassan et al. 2013; Sharif et al. 2014; Zimbra et al. 2018), there has been limited work on constructs like trust, anxiety, and perceptions of literacy.

- Given that psychometric analysis often entails user modeling that could involve analysis of text, survey-based responses (psychometric construct measures), and demographics, our test bed encompasses all three types of data (Abbasi et al. 2015).

- For each user, we capture text and gold-standard survey responses for four psychometric dimensions. The combination of four target dimensions, coupled with the aforementioned demographic and additional survey data affords opportunities for advanced machine learning text classification approaches such as multi-task learning, and psychometric embeddings and encoders (Ahmad et al. 2020).

- By including text and demographics from diverse user populations, the testbed also presents interesting opportunities for text machine learning research on fairness in NLP models (Abbasi et al. 2018; Taylor et al. 2018).

- While our efforts are geared towards psychometric dimensions in the health context, the method employed can be generalized to various contexts where psychometric dimensions are possible, practical, and valuable.

As noted, we believe the testbed and process have important implications for future NLP research that examines psychometrics as part of broader user modeling efforts. In the rest of the paper we describe the testbed construction process, summary statistics, and some preliminary results on psychometric classification tasks.

## 2. Related Work – Psychometric Language Resources

Over the past thirty years, significant efforts have been made to develop a robust and burgeoning set of language resources for various linguistic and NLP tasks. Gold-standard enriched test beds have been developed for sentiment analysis, including sentiment polarity (e.g., positive, neutral, negative), sentiment aspects, targets, opinion holder analysis, and so on (Wiebe et al. 2005; Thelwall et al. 2010; Sanders 2011). Similarly, testbeds have been developed for affect and emotion. Personality traits manifested in test have also received attention (Luyckx and Daelemans 2008). More recent work has explored construction of corpora for examining depression and cyberbullying, with the latter also annotating self-disclosures of personal information which may trigger bullying (Rakib and Soon 2018).

Given that psychometrics is concerned with measurement of attitudes, beliefs, perceptions, and personality traits, many of these aforementioned test beds and avenues of language resource construction could be considered as focusing on psychometric dimensions (Ahmad et al. 2020). As noted in the introduction section, our work builds on this work by focusing on underexplored dimensions such as trust, anxiety, and perceptions of literacy. Moreover, rather than relying on independent annotation, we seek to utilize user-generated text that is captured along with self-reported survey-based responses for the psychometric dimensions of interest. Hence, the text is accompanied by survey-based quantifications from the individuals that can serve as a gold-standard proxy of what we hope to measure/score by applying NLP methods to the users' text. This paper explores the efficacy of such a method that bridges the social science and NLP perspectives for test bed construction. We later use supervised machine learning classification methods to demonstrate the viability of the approach – that is, to validate that the text samples captured can indeed serve as a reasonable proxy of the users' survey-based responses for the psychometric dimensions of interest. Further, our testbed also includes the users' survey-based responses to related psychometric dimensions, as well as demographic data. Details are as follows.

## 3. Testbed Construction Process

In this section we describe the process taken to construct our psychometric NLP test bed. The key steps included identifying relevant psychometric dimensions of interest, finding suitable survey-based items to operationalize our latent constructs, assessing different prompts for text equivalency questions, and test bed construction validation. Details regarding these steps are as follows.

## 3.1 Identifying Key Psychometric Dimensions

Given our focus on psychometrics in the healthcare context, we began by reviewing nearly 90 articles from the behavioral health literature (refs). These articles all used survey-based methods to measure a set of core psychometric dimensions (i.e., latent constructs). Based on our literature review, we developed a structural equation model (see Figure 2) that showed the relevant antecedent-consequent relations between various psychometric dimensions (Netemeyer et al. 2019). A structural equation model is a multivariate statistical analysis technique that is used to analyze structural relationships between measured variables and latent constructs. As depicted in Figure 2, the ovals represent psychometric dimensions and the arrows denote relationships. For instance, based on the literature, one's perceptions of their health literacy are expected to influence their trust in physicians and anxiety visiting the doctor. Similarly, trust in physicians impacts future doctor visits and all-around well-being. In order to validate the initial identified set of psychometric dimensions, we ran the structural equation model against collected survey-based data. The values along the arrows indicate effect sizes for significant paths observed in different population segments. Insignificant paths are denoted with "ns". With the exception of the link from anxiety visiting the doctor to future doctor visits, all paths were significant. These results suggested that the identified psychometric dimensions of interest might be worth exploring further.

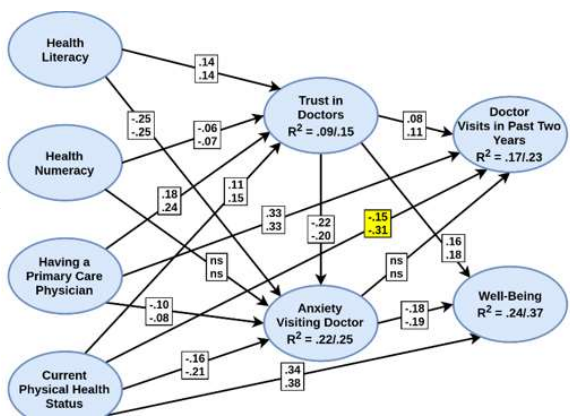

**Figure 2**: Structural equation model depicting key psychometric dimensions (and their relations), observed in the literature.

## 3.2 Developing Survey Items

Based on the analysis described in section 3.1, we further narrowed the consideration set down to four psychometric dimensions based on suitability of text-based response collection: trust in physicians, anxiety visiting the doctor, subjective literacy, and objective health numeracy. A critical step in survey-based psychometric research performed in the social sciences is development or inclusion of appropriate items to measure the latent constructs. Through our review of the literature, our own survey-based data collection, and statistical analysis (exploratory and confirmatory factor analysis), we identified a subset of items for each of these dimensions. An overview of the four psychometrics dimensions and some of their related items is as follows:

*Health Literacy (HL)* – In essence, HL is a subjective construct reflecting how much one thinks one knows about health and access to health-related information and providers (Osborne et al. 2013). Low HL has been associated with increased mortality, increased hospitalization, and poor adherence and self-maintenance to a host of chronic diseases such as diabetes, heart disease, and risk of stroke (Atlin et al. 2014; Berkman et al. 2011; Osborne et al. 2013). Low HL has also been shown to be more prevalent among the elderly, lower income and education groups, and certain racial groups (Atlin et al. 2014). In total, 10 HL items from three different scales were incorporated. Figure 3 shows examples of three of the items incorporated, which relate to one's perceptions of ability to understand hospital materials, process medical information, and comprehend medical conditions.

*Health Numeracy (HN)* – Conversely, health numeracy (HN) is an objective construct reflecting the ability to calculate, use, and understand numeric and quantitative concepts in the context of health issues (Schapira et al. 2014). HN has been associated with positive health outcomes such as the ability to understand dosage in medication and adherence to self-care diabetes treatment (Ciampa et al. 2010; Osborn et al. 2013). As with HL, lower HN scores are more prevalent among the elderly, lower income and education groups, and certain racial groups (Schapira et al. 2014). We incorporated two HN scales comprising 14 total items. Figure 4 depicts four item examples from one of the two scales utilized. As shown, these items are objective measures such as ability to count calories or read a thermometer.

*Trust in Doctors (TD)* – Perceptions of trust in physicians/doctors (TD) can have an important mediating role on health outcomes (Dugan et al. 2005). TD was measured using the 5 items depicted in Figure 1.

*Anxiety Visiting Doctors (AV)* – Anxiety when visiting the doctor's office is another strong potential mediator for health outcomes such as future doctor visits and wellness (Spielberger 1989; Marteau and Bekker 1992). Figure 5 shows the items used to measure AV. These focused on levels of anxiousness, worry, uncertainty, and uneasiness.

## 3.3 Attaining User-Generated Text

We used an iterative trial-and-error process to develop our "equivalent" user generated text related to the four aforementioned psychometric dimensions. After several rounds of face validity checks and piloting with small sets of respondents, we ultimately arrived at a configuration where the survey items were used to prime respondents. We immediately followed these items with text questions that were tuned as part of our iterative process.

**Q1** How often do you have someone help you read hospital materials?
- Always (1)
- Often (2)
- Sometimes (3)
- Occasionally (4)
- Never (5)

**Q2** How often do you have problems learning about your medical condition because of difficulty understanding written information?
- Always (1)
- Often (2)
- Sometimes (3)
- Occasionally (4)
- Never (5)

**Q3** How often do you have a problem understanding what is told to you about your medical condition?
- Always (1)
- Often (2)
- Sometimes (3)
- Occasionally (4)
- Never (5)

**Figure 3**: Examples of survey items related to subjective health literacy (HL).

**MN1** James has diabetes. His goal is to have his blood sugar between 80 mg/dL and 150 mg/dL in the morning. Which of the following blood sugar readings is within his goal?
- 55 mg/dL (1)
- 140 mg/dL (2)
- 165 mg/dL (3)
- 180 mg/dL (4)

**MN2** Nathan has a pain rating of 5 on a pain scale of 1 (no pain) to 10 (worst possible pain). One day later Nathan still has pain but not as much. Now, what pain rating might Nathan give?
- 3 (1)
- 5 (2)
- 7 (3)
- 9 (4)

**MN8** A nutrition label is shown below. How many calories did Mary eat if she had 2 cups of food?
- 140 calories (1)
- 280 calories (2)
- 560 calories (3)
- 680 calories (4)

Graphic

| Nutrition Facts | |
|---|---|
| Serving Size 1 cup (228g) | |
| Servings per Container 2 | |
| Amount Per Serving | |
| Calories 280 | Calories from Fat 120 |
| | % Daily Value* |
| Total Fat 13g | 20% |
| Saturated Fat 5g | 25% |
| Trans Fat 2g | |
| Cholesterol 2mg | 10% |
| Sodium 660 mg | 28% |
| Total Carbohydrate 31g | 10% |
| Dietary Fiber 3g | |
| Sugars 5g | |
| Protein 5g | |
| Vitamin A 4% | Vitamin C 2% |
| Calcium 15% | Iron 4% |
| *Percent Daily Values are based on a 2,000-calorie diet. Your Daily values may be higher or lower depending on your calorie needs | |

**Figure 4**: Examples of survey items related to objective health numeracy (HL).

Below are some emotions that might be used to describe how you feel when having a doctor give you a health examination. Please think about each emotion carefully and whether a physician health examination made you feel...

(Scale: Strongly Disagree (1), Disagree (2), Somewhat Disagree (3), Neither Agree nor Disagree (4), Somewhat Agree (5), Agree (6), Strongly Agree (7))

- Anxious
- Upset
- Discouraged
- Fearful
- Worried
- Uneasy
- Dread
- Uncertainty

**Figure 5**: Examples of survey items related to anxiety in visiting the doctor (AV).

The text-response questions yielded the best responses when the questions were at the end of the survey item section for that particular psychometric dimensions, appearing immediately at the bottom of the same/final page of survey items. Figure 6 illustrates the method for collecting text responses after survey-based priming. Table 1 depicts the prompts or questions used to attain the user-generated text responses. While we recognize that the questions asked and approach undertaken could be further enhanced, we believe this constitutes an important first step toward aligning survey items with user-generated text responses. As we later show in the evaluation section, preliminary results from text classification tasks lend validity to the construction.

Patient Trust in a Physician

(Scale: Strongly Disagree (1), Disagree (2), Neutral (3), Agree (4), Strongly Agree (5))

- Sometimes your doctor cares more about what is convenient for (him/her) than about your medical needs.
- Your doctor is extremely thorough and careful.
- You completely trust your doctor's decisions about which medical treatments are best for you.
- Your doctor is totally honest in telling you about all of the different treatment options available for your condition.
- All in all, you have complete trust in your doctor.

In a few sentences, please explain the reasons why you trust or distrust your primary care physician. If you do not have a primary care physician, please answer in regard to doctors in general.

**Figure 6**: Illustration of survey item-primed text response collection (example shows trust in physician TD).

| Psychometric Dimension | Question or Prompt |
|---|---|
| Anxiety visiting the doctor (AV) | In a few sentences, please describe what makes you most anxious or worried visiting the doctor's office |
| Subjective health literacy (HL) | Regarding all the questions you just answered, to what degree do you feel you have capacity to obtain, process, and understand basic health information and services needed to make appropriate health decisions? Please explain you answer in a few sentences. |
| Trust in physicians (TD) | In a few sentences, please explain the reasons why you trust or distrust your primary care physician. If you do not have a primary care physician, please answer in regard to doctors in general. |
| Objective health numeracy (HN) | In a few sentences, please describe an experience in your life that demonstrated your knowledge of health or medical issues. |

**Table 1:** Questions used to elicit user-generated text responses

## 4. Testbed Results and Summary Statistics

Two rounds of data collection were performed. In the first round, we collected data using Amazon Mechanical Turk (AMT). A total of 4,262 usable responses were attained through AMT. In order to attain a second, more diverse set of responses, Qualtrics was used to collect an additional 4,240 clean responses. Table 2 shows the testbed summary statistics. Each respondent provided a text response for each of the four psychometric dimensions, in addition to survey responses to all dimension items as well as additional demographic and behavior questions and complementary psychometric dimensions. In other words, we received 17,048 total responses from AMT (4262 x 4) and 16,960 from Qualtrics (4240 x 4), total. The AMT respondents tended to be more representative of the overall US population in terms of race, gender, and education. As noted earlier, the goal of the Qualtrics data collection was to garner a richer sample of responses from diverse populations with greater representation for racial and gender minorities and education and income-based disparate populations. Such testbeds are important to allow deeper exposition into studies that examine fairness of NLP models (Abbasi et al. 2018).

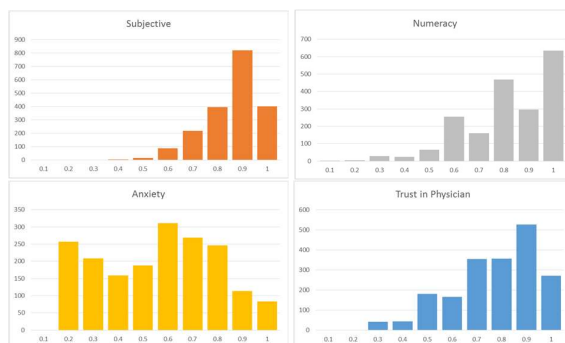

**Figure 7**: Distribution of mean survey-based response scores for psychometric dimensions on the AMT test bed.

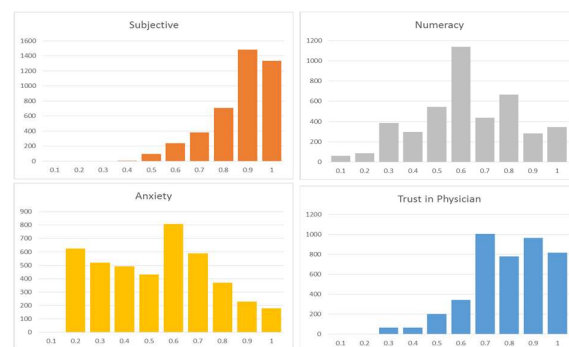

**Figure 8**: Distribution of mean survey-based response scores for psychometric dimensions on the Qualtrics data.

Table 3 shows examples of psychometric scores and accompanying text responses for the subjective health literacy (HL) dimension. The scores were scaled from 0-1 based on the survey responses. The accompanying user text responses correspond to the two users' self-reported scores. The example illustrates the "alignment-oriented" objectives of testbed construction in this context.

| Characteristics | AMT | Qualtrics |
|---|---|---|
| User Responses | 4,262 | 4,240 |
| Text Fields (per user respondent) | Subjective literacy (SL) Objective numeracy (HN) Trust in physicians (TD) Anxiety visiting the doctor (AV) | |
| Race | 81.2% white 7.4 % black | 50% white 50% black |
| Gender (male) | 48.3% | 24.2% |
| Income (USD) | 62% < $55K | 67% < $55K |
| Education (college grads) | 44.6% | 32.1% |
| Examples of other behavior/ psychometric dimensions | Usage of prescription drugs Presence of primary care physician Frequency of doctor visits Smoking and drinking frequency | |

**Table 2:** Summary of the AMT and Qualtrics psychometric NLP test beds

The most critical survey response items in the data are the ones corresponding to the four dimensions. Following best practices in the social science literature on survey design, we constructed a single composite score for each of these dimensions by averaging across all underlying items. The scores were scaled to a 0-1 range. Figure 7 and 8 depict the distribution of user responses for the four dimensions (subjective HL, numeracy HN, trust TD, and anxiety AV) on the AMT and Qualtrics data collections. We can see that for subjective, numeracy, and trust the response followed a skewed Gaussian distribution. In contrast, the distribution for anxiety scores was more uniformly distributed. Further, as expected, there were differences between the AMT and Qualtrics populations. For instance, AMT respondents had higher numeracy and trust in physicians. Conversely, the two data collections were relatively more comparable on the subjective literacy and anxiety visiting the doctor dimensions.

| HL score | Text Response for Subjective Literacy Prompt |
|---|---|
| 0.4667 | I feel like with the terms and complicated medical lingo, I am not exactly sure what some of the meanings entail. Such as If I am diagnosed with a certain condition and need medication X, I don't know what that medication does, what the alternatives are, I don't even know how to pronounce some of these names. I feel like I am able to ask the doctors but can not fully grasp the magnitude of the situation without looking at the whole picture which is difficult to have someone explain to me in one visit. I feel like I need a step by step diagnostic, of why this happened, what I can do, what the consequences are, what some definitions of the disease are, etc. |
| 0.9167 | I think I have a fine capacity. I am able to coherently explain my concerns, and ask for aid if I need it. I am native in English, and know all my health issues and past surgeries and such. It isn't hard for me to do anything medical, and I am confident in making whatever medical decisions I need to make. |

**Table 3:** Examples of survey-based scores and accompanying text responses for subjective health literacy

## 5. Testbed Construction Evaluation

In order to evaluate the effectiveness of the constructed psychometric NLP data set, we conducted a series of text classification experiments to see how well machine learning classification models could predict survey-based "gold-standard" ratings (treated as the dependent variable in our context) using the free text responses. A standard five-fold cross-validation scheme was employed where in each fold, a different 20% of the data was set aside for testing while the remaining 80% was used for training.

In order to convert our continuous survey-based dependent variables to binary classification labels, we bifurcated the data by using the bottom 25% of data points as the low value and the top 25% as the high value. For instance, the top quartile of trust in physician values were deemed "high trust" whereas the bottom quartile were "low trust." We used a support vector machines (SVM) classifier coupled with a classic feature subsumption-based feature selection method (Riloff et al. 2006) called feature relation network (FRN) (Abbasi et al. 2011). FRN uses rule-based feature selection applied to a large set of word, part-of-speech, character, and domain lexicon-based n-grams to identify a subset of key predictors. This method was used since it has attained reasonable results on related text classification tasks such as sentiment analysis as a feature-based classification method. We have attained even stronger results using more advanced deep learning methods in our more recent work (Ahmad et al. 2020).

Table 4 presents the 5-fold cross validation results on the Qualtrics subset of the testbed. Similar results were attained on AMT as well. The table shows the accuracy, precision, recall, and f1 measure for the best FRN feature quantity as well as the results when fixing the number of input features to 15,000. We can see that for the trust (TD), subjective literacy (HL), and numeracy (HN) tasks, the classifiers attained around 80% accuracy. This is on par with many other classification tasks such as binary sentiment classification on Twitter (Abbasi et al. 2014). We also included the number of features utilized since the quantity of n-grams yielding the best performance can be considered a measure of the linguistic variation incorporated by users – larger feature set sizes are sometimes indicative of a more complex linguistic phenomena where the classifier needs to account for "long tail" usage of terms and literary devices. Here that does not appear to be the case. The one obvious exception is the anxiety task, where the best classifier only garnered about 73% accuracy on the balanced data set.

Figure 9 shows the impact of feature quantity on classification accuracy across the four tasks. As noted, the quantity of features needed to attain best performance as well as the stability of classification performance are both indicators of the viability of the test classification task (Abbasi et al. 2011). The results underscore the efficacy of the survey-text alignment.

| **Trust in Physician** | | | | | |
|---|---|---|---|---|---|
| Feature Set | Accuracy | Precision | Recall | F1 | No. Features |
| FRN Best | 80.99 | 81.12 | 80.85 | 80.97 | 40,000 |
| FRN 15K | 80.10 | 80.48 | 79.65 | 80.00 | 15,000 |
| **Health Anxiety** | | | | | |
| Feature Set | Accuracy | Precision | Recall | F1 | No. Features |
| FRN Best | 73.21 | 72.72 | 74.34 | 73.47 | 20,000 |
| FRN 15K | 72.92 | 72.92 | 73.04 | 72.95 | 15,000 |
| **Subjective Literacy** | | | | | |
| Feature Set | Accuracy | Precision | Recall | F1 | No. Features |
| FRN Best | 79.01 | 79.12 | 78.88 | 78.98 | 40,000 |
| FRN 15K | 78.63 | 78.54 | 78.98 | 78.71 | 15,000 |
| **Health Numeracy** | | | | | |
| Feature Set | Accuracy | Precision | Recall | F1 | No. Features |
| FRN Best | 79.99 | 79.66 | 80.55 | 80.09 | 50,000 |
| FRN 15K | 79.10 | 79.12 | 79.14 | 79.11 | 15,000 |

**Table 4:** Classification performance for FRN-based SVM classifier on the Qualtrics subset of the testbed.

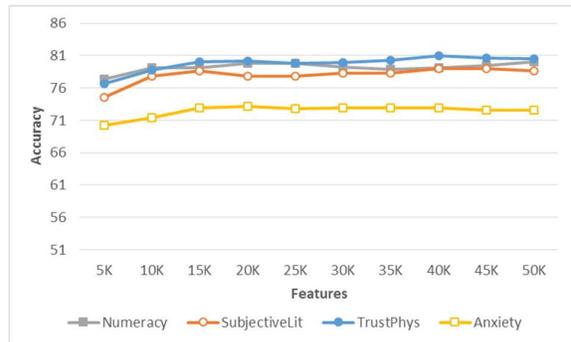

**Figure 9**: Impact of quantity of features in FRN-based SVM classifiers on classification performance across the four psychometric tasks in Qualtrics data.

## 6. Conclusion

The results of our work have important implications for several stakeholder groups. NLP research focused on constructing novel empirical methods can use the constructed test bed to build new text analytics models for psychometric NLP. The inclusion of demographic, text, target psychometric, and secondary psychometric data in the testbed could allow development of rich deep learning architectures that incorporate user models, psychometric embeddings, structural equation model-based encoders, and multi-task learning across the four parallel target psychometric dimensions. The unique multimodal nature of the data may also afford opportunities to better understand and study fairness in NLP models and methods. For each text utterance, the testbed encompasses gender, race, education levels, and income – all fields that are often the basis for bias in machine learning algorithms. The examination of fairness in NLP is an underexplored area of research. Finally, other teams developing language resources can adapt the process outlined to other domains such as security, e-commerce, finance, etc. We recognize that this is a first foray into rich psychometric NLP. Our hope is that future work can improve upon the methods and best practices for examining the interplay between survey-based constructs and their manifestations in user-generated text.

## 7. Acknowledgements

We would like to acknowledge support of this work through the following grants: NSF BDS-1636933; NSF IIS-1816504; NSF IIS-1553109 and Microsoft Research CRM:0740129.